\documentclass[letterpaper, 10 pt, conference]{ieeeconf}  

\IEEEoverridecommandlockouts                              

\usepackage{graphicx}
\usepackage{amsmath} 
\usepackage{amssymb}  

\usepackage{subcaption}
\usepackage{array}
\usepackage{flushend}
\usepackage{cite}   
\usepackage[dvipsnames]{xcolor}
\usepackage{pifont}

\newcommand{\xmark}{\ding{55}}

\usepackage{hyperref}

\hypersetup{
    colorlinks=true,
    linkcolor=black,
    urlcolor=cyan,
}

\usepackage{amsmath, amssymb}

\usepackage{xifthen}
\usepackage{color}

\newcommand{\ba}{\begin{eqnarray}}
\newcommand{\ea}{\end{eqnarray}}

\newcommand{\unit}[1]{\mbox{$\mathrm{\,#1}$}}

\renewcommand{\deg}{\mbox{${}^\circ$}}

\newcommand{\R}{\mathbb{R}}

\newcommand{\presup}[1]{\,{}^{\scriptscriptstyle #1}\!}

\newcommand{\pose}[1][ZZZZ]{\ifthenelse{\equal{#1}{ZZZZ}}{}{\presup{#1}}{\mathbf{\xi}}}
\newcommand{\estpose}[1][ZZZZ]{\ifthenelse{\equal{#1}{ZZZZ}}{}{\presup{#1}}{\mathbf{\hat{\xi}}}}
\newcommand{\hpose}[1][ZZZZ]{\ifthenelse{\equal{#1}{ZZZZ}}{}{\presup{#1}}{\hat{\mathbf{\xi}}}}
\newcommand{\posedot}[1][ZZZZ]{\ifthenelse{\equal{#1}{ZZZZ}}{}{\presup{#1}}{\mathbf{\nu}}}
\newcommand{\q}[1][ZZZZ]{\ifthenelse{\equal{#1}{ZZZZ}}{}{\presup{#1}}{\mathring{q}}}
\DeclareMathAlphabet{\mathitbf}{OML}{cmm}{b}{it}
\newcommand{\twist}[2][ZZZZ]{\ifthenelse{\equal{#1}{ZZZZ}}{}{\presup{#1}}{\mathcal{S}}}
\renewcommand{\vec}[2][ZZZZ]{\ifthenelse{\equal{#1}{ZZZZ}}{}{\presup{#1}}{\mathitbf{#2}}}
\newcommand{\hvec}[2][ZZZZ]{\ifthenelse{\equal{#1}{ZZZZ}}{}{\presup{#1}}{\tilde{\vec{#2}}}}
\newcommand{\evec}[2][ZZZZ]{\ifthenelse{\equal{#1}{ZZZZ}}{}{\presup{#1}}{\hat{\vec{#2}}}}
\newcommand{\bvec}[2][ZZZZ]{\ifthenelse{\equal{#1}{ZZZZ}}{}{\presup{#1}}{\bar{\vec{#2}}}}
\newcommand{\dvec}[2][ZZZZ]{\ifthenelse{\equal{#1}{ZZZZ}}{}{\presup{#1}}{\dot{\vec{#2}}}}
\newcommand{\ddvec}[2][ZZZZ]{\ifthenelse{\equal{#1}{ZZZZ}}{}{\presup{#1}}{\ddot{\vec{#2}}}}

\newcommand{\mat}[2][ZZZZ]{\ifthenelse{\equal{#1}{ZZZZ}}{}{\presup{#1}\,}{{\boldsymbol #2}}}
\newcommand{\dmat}[2][ZZZZ]{\ifthenelse{\equal{#1}{ZZZZ}}{}{\presup{#1}\,}{{\dot{\boldsymbol #2}}}}
\newcommand{\emat}[2][ZZZZ]{\ifthenelse{\equal{#1}{ZZZZ}}{}{\presup{#1}\,}{\hat{\boldsymbol#2}}}
\newcommand{\matfn}[3][ZZZZ]{\ifthenelse{\equal{#1}{ZZZZ}}{}{\presup{#1}}{{\mat{#2}}\left(#3\right)}}
\newcommand{\Rt}[2][ZZZZ]{\ifthenelse{\equal{#1}{ZZZZ}}{}{\presup{#1}}{{\bf R}\left(#2\right)}}

\newcommand{\point}[2][ZZZZ]{\ifthenelse{\equal{#1}{ZZZZ}}{}{\presup{#1}}{\mathbf{\mathrm{#2}}}}

\newfont{\School}{pncr}
\newfont{\eightTR}{pncr at 8pt}

\usepackage{color}
\usepackage{fancyvrb}
\fvset{formatcom=\color{blue},fontseries=c,fontfamily=courier,xleftmargin=4mm,commentchar=!}
\DefineVerbatimEnvironment{Code}{Verbatim}{formatcom=\color{blue},fontseries=c,fontfamily=courier,fontsize=\footnotesize,xleftmargin=4mm,commentchar=!}
\DefineVerbatimEnvironment{CodeSmall}{Verbatim}{formatcom=\color{blue},fontseries=c,fontfamily=courier,fontsize=\scriptsize,xleftmargin=1mm,commentchar=!}
\DefineVerbatimEnvironment{CodeNum}{Verbatim}{numbers=left,numbersep=4pt,formatcom=\color{blue},fontseries=c,fontfamily=courier,fontsize=\footnotesize,xleftmargin=4mm}

\newcommand{\model}[1]{\index{code}{#1@\textit{#1}}\ifthenelse{\boolean{draft}}{{\color{green}\Verb+#1+}}{\Verb+#1+}}
\newcommand{\block}[1]{\ifthenelse{\boolean{draft}}{{\color{green}\Verb+#1+}}{\textsf{#1}}}
\newcommand{\func}[2][ZZZZ]{\ifthenelse{\equal{#1}{ZZZZ}}{\index{code}{#2}}{\index{code}{#1}}\ifthenelse{\boolean{draft}}{{\color{green}\Verb+#2+}}{\Verb+#2+}}
\newcommand{\methodb}[2]{\index{code}{#1@\textbf{#1}!.#2}\ifthenelse{\boolean{draft}}{{\color{magenta}\Verb+#1.#2+}}{\Verb+#1.#2+}}
\newcommand{\method}[2]{\index{code}{#1@\textbf{#1}!.#2}\ifthenelse{\boolean{draft}}{{\color{magenta}\Verb+#2+}}{\Verb+#2+}}
\newcommand{\class}[1]{\index{code}{#1@\textbf{#1}}\ifthenelse{\boolean{draft}}{{\color{cyan}\Verb+#1+}}{\Verb+#1+}}
\newcommand{\property}[1]{\index{property}{#1}\ifthenelse{\boolean{draft}}{{\color{cyan}\Verb+#1+}}{\Verb+#1+}}

\newcommand{\SE}[1]{\ensuremath{\mathrm{{\bf SE}(#1)}}}

\makeatletter
\newcommand*{\balancecolsandclearpage}{%
  \close@column@grid
  \clearpage
  \twocolumngrid
}
\makeatother

\def\showcomments{1}

\usepackage{xcolor}
\newcommand{\peter}[1]{{
    \if\showcomments1
        \color{red}#1
    \fi
}}
\newcommand{\jesse}[1]{{
    \if\showcomments1
        \color{blue}#1
    \fi
}}
\newcommand{\feras}[1]{{
    \if\showcomments1
        \color{brown}#1
    \fi
}}

\title{\LARGE \bf
A Purely-Reactive Manipulability-Maximising Motion Controller}

\author{Jesse Haviland$^{1}$, Peter Corke$^{1}$
\thanks{$^{1}$Jesse Haviland and Peter Corke are with the Australian Centre for Robotic Vision (ACRV), Queensland University of Technology Centre for Robotics (QCR), Brisbane, Australia
        {\tt\small j.haviland@qut.edu.au, peter.corke@qut.edu.au}. This research was conducted by the Australian Research Council project number CE140100016, and supported by the QUT Centre for Robotics.
}%
}

\begin{document}

\maketitle
\thispagestyle{empty}
\pagestyle{empty}

\begin{abstract}
We present a novel approach to controlling the instantaneous velocity of a robot end-effector that is able
to simultaneously maximise manipulability and avoid joint limits.
It operates on non-redundant and redundant robots, which is achieved by adding artificial
redundancy in the form of controlled path deviation.
We formulate the problem as a quadratic programme and provide an open-source Python
implementation that provides solutions in just a few milliseconds.
It accepts a robot model expressed using URDF or Denavit-Hartenberg
parameterisation.
We compare our method to previous work in simulation and on a physical robot. 
\end{abstract}


\section{Introduction}

Resolved-rate motion control (RRMC)\cite{rrmc} is an old (1969) but effective technique to control
the velocity of a robot's end-effector in task space, and it is readily applied to redundant
manipulators.
The ability of a robot manipulator to achieve an arbitrary end-effector velocity is a function of the manipulator
Jacobian and this can be summarised as a scalar such as the manipulabilty measure proposed by \cite{manip} in 1985.

Choosing the end-effector velocity to achieve the task can be achieved using 
 techniques such as potential fields \cite{re0}, resolved-rate motion control \cite{rrmc}, or quadratic programming \cite{qp1}.
These approaches can be augmented to achieve secondary tasks such as joint-limit avoidance \cite{re0} 
and manipulability maximization \cite{qp0}.
For example \cite{gpm} achieves manipulability maximization by projecting the gradient of the manipulability into the null space of the differential kinematics.
The ability to achieve a secondary task has previously only been possible if the robot is redundant, by exploiting null-space
motion.

Despite the obvious utility of velocity control and joint configuration optimised for manipulability 
and joint-limit avoidance, and the existence of these techniques for decades, it is
surprising that they are not in everyday use.
We observe that many implementers persist with packages such as MoveIt! \cite{moveit} which is slow and often produces (perhaps due to lack of understanding by the users) quite unsatisfactory paths.  We speculate that the algorithms mentioned earlier are either not widely known, or that there are no software tools available which leads users to go with what is available rather than what is best for the task.

\begin{figure}[t]
    \centering
    \includegraphics[height=4.9cm]{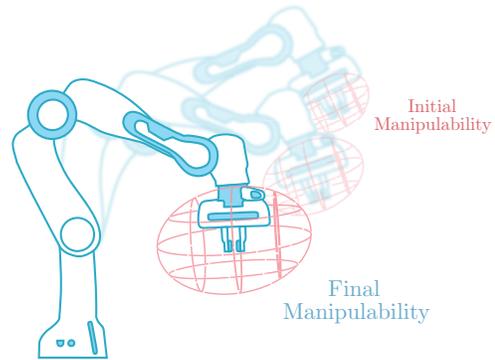}
    \caption{
        Our Manipulability Motion Controller drives a robot's joint velocities such that the manipulability of the robot is maximised, while achieving the desired end-effector pose and maintaining joint limits. The manipulability can be visualised through velocity ellipsoids created using the manipulator Jacobian (more details in Figure \ref{fig:large}).
    }
    \label{fig:cover}
\end{figure}

In this paper we describe a set of tools for controlling robot end-effector velocity  that optimizes multiple objectives.
Providing control of end-effector velocity allows for reactive control such as visual servoing and closed-loop grasping.
We incorporate extra redundancy into the task by allowing controlled deviation along the path, but
not at the goal, to avoid joint limits and maximise manipulability.
Critical to optimizing manipulability is computing the gradient of the manipulability measure with respect to joint velocity, which involves the manipulator Hessian.  Unlike the prior art, we provide tools to compute this quickly for any manipulator, even those that are not describable using Denavit-Hartenberg notation.

The contributions of this paper are:
\begin{enumerate}
    \item a control framework for reactive multi-objective end-effector velocity control.
    This works for redundant and non-redundant
    and for robots whose kinematics is not neccessarily described in Denavit-Hartenberg notation, for example URDF format. We use slack variables to accommodate conflict between objectives.
    \item experimental validation in simulation with several popular serial link manipulators, comparison with existing techniques, and experimental validation on a physical Franka-Emika Panda robot arm.
    \href{https://jhavl.github.io/mmc}{jhavl.github.io/mmc}.
\end{enumerate}

\newcolumntype{M}[1]{>{\centering\arraybackslash}m{#1}}
\begin{table*}[t]
    \centering
    \renewcommand{\arraystretch}{1.3}

    \caption{Features Comparison of Reactive Motion Controllers incorporating Manipulability}
    \label{tab:comp}

    \begin{tabular}{ c | M{1.8cm} | M{1.8cm} | M{1.8cm} | M{1.8cm} | M{1.8cm} | M{1.8cm} | M{1.8cm}}
    \hline
    {Controller} & {DH Robot Models \cite{dh}} & {URDF Robot Models \cite{urdf}} & {ETS Robot Models \cite{ets2, ets}} & {Avoid Joint Position Limits} & {Avoid Joint Velocity Limits} & {Non-redundant Robots} & {Code Avaliable} \\
    \hline\hline
    MMC (ours)  & \color{ForestGreen} \checkmark & \color{ForestGreen} \checkmark & \color{ForestGreen} \checkmark & \color{ForestGreen} \checkmark & \color{ForestGreen} \checkmark & \color{ForestGreen} \checkmark & \color{ForestGreen} \checkmark \\
    Park \cite{gpm}  & \color{ForestGreen} \checkmark & \color{Red} \xmark & \color{Red} \xmark & \color{Red} \xmark & \color{Red} \xmark & \color{Red} \xmark & \color{Red} \xmark \\
    Baur \cite{gpm2}  & \color{ForestGreen} \checkmark & \color{Red} \xmark & \color{Red} \xmark & \color{ForestGreen} \checkmark & \color{Red} \xmark & \color{Red} \xmark & \color{Red} \xmark \\
    \hline
    \end{tabular}
\end{table*}

\section{Related Work}\label{sec:related_work}

The goal, when kinematically controlling serial-link manipulator, is to find a control which provides the desired end-effector motion in the manipulator's task space. 
There is a linear mapping between the instantaneous end-effector spatial velocity and the joint velocities given by the manipulator Jacobian which is a $6 \times n$ matrix and where $n$ is the number of robot joints.
For a robot where $n=6$ the Jacobian is square and if invertible can be used to map task space velocity to joint space velocity. This technique is the standard approach for reactive kinematic control in the velocity domain and is known as resolved-rate motion control \cite{rrmc}.

However, if $n>6$, the manipulator Jacobian is not square and consequently, the inverse is not possible. There are an infinite number of joint velocity vectors which will give rise to the task-space velocity. We commonly use the Moore-Penrose pseudoinverse which yields the joint space velocity with the minimum Euclidean norm. 

If the condition number of the Jacobian is high, or the Jacobian is rank deficient, then some task-space velocities are unachievable or only achievable with very high joint velocities \cite{peter}. This has led to several approaches which use optimisation strategies for redundant manipulators.

Quadratic programs are a powerful tool for optimisation which can represent complex systems while always being solvable in a finite time (or shown to be infeasible) \cite{opt0}. Quadratic programming, in general, can incorporate equality, inequality, and bound constraints simultaneously. The pseudoinverse can be modelled as a quadratic programming problem. In contrast to non-linear programming, the objective function used in a quadratic programming problem is convex (under certain conditions, see Section \ref{sec:qp})\cite{opt0}. Therefore, a unique solution exists and can always be found. From a quadratic programming perspective, the pseudoinverse solution minimises the control input, in terms of joint velocity. However, this solution does nothing to stop the robot from reaching a singular configuration.

More recent work on kinematic control of redundant manipulators uses a planning-based paradigm \cite{mp0, qp2}. In motion planning, joint motion is generated for the entire movement from the robot's starting pose to the goal. Recent progress in this area has seen the kinematic motion planning problem solved using techniques such as quadratic programming and neural networks. 
In \cite{qp1}, quadratic programming was used to aid in obstacle avoidance with redundant manipulators, while in \cite{qp0} it was used to maximise manipulability.

Neural networks have also been utilised for motion planning with redundant manipulators. The work in \cite{nn0} used a dynamic neural network to choose joint velocities which increase manipulability, while also staying within the physical joint velocity limits of the robot. This is similar to the work in \cite{nn1, nn2, nn4, nn5}, however, the controllers devised in these works do not optimise for manipulability. Alternatively, a learning based approach was devised which allowed a robot to reproduce manipulability ellipsoids, essentially maximising manipulability in certain directions \cite{geo0, geo1}.
Recent work \cite{qp3} incorporated the physical joint limits of a mobile manipulator into a quadratic programming function. However, it does not assist the robot in avoiding singularities or maximising manipulability.

Motion planning solutions are able to compute optimal paths based on global knowledge of the goal and
constraints, but they do not provide the level of reactivity required for control techniques such as visual servoing \cite{vs0} and closed-loop visual grasping \cite{doug}. Purely reactive controllers allow arbitrary end-effector velocity to be set at each iteration of the sensor-based control loop.  At each time step, the controller must make the best choice of joint velocity that meets several instantaneous constraints including the current joint configuration which is the integral of previous control decisions -- this admits the
possibility of failure.

Maximizing manipulability is particularly critical for reactive control which must be open to a velocity
demand in any task-space DoF.
Manipulability maximization was first achieved by Park \cite{gpm} where the gradient of the manipulability was projected into the null-space of the differential kinematics. This approach, while requiring no optimisation, only works on redundant robots. Furthermore, the proposed solution was formulated for robots modelled with Denavit-Hartenberg (DH) parameters \cite{dh}. Whilst most existing industrial robots can be represented using DH parameters, modern robots are modelled using universal robot description format (URDF) \cite{urdf}.  URDF and elementary transform sequence (ETS) \cite{ets2, ets} can model any robot. The work in \cite{gpm3} used the gradient projection method to assist with human-robot cooperation, while \cite{gpm4} used it for redundancy resolution on a mobile manipulator. Baur \cite{gpm2} expanded \cite{gpm} to provide a null-space projection which can improve manipulability while also encouraging joint-position limits to be respected on an agricultural manipulator using the velocity damper approach \cite{pp0}.

These approaches, and RRMC, drive the robot's end-effector along a straight line towards the goal pose but
this can also be their undoing. If the robot encounters a singularity, or a joint limit along this path, it will be unable to avoid it and fail. Hence, in this paper we introduce extra redundancy to the system via intentional error to the end-effector path, called slack \cite{opt1}. Through slack, we can avoid the aforementioned issues, while also allowing our controller to operate on non-redundant robots. As shown in Table \ref{tab:comp}, we provide significant functionality compared to the state-of-the-art and also
provide the tools to implement our controller on any robot using the open-source Robotics Toolbox for Python \cite{rtb}.

In Section \ref{sec:rrmc} we outline the traditional approach to resolved-rate motion and then relate this to a quadratic programming problem in \ref{sec:qp}. Section \ref{sec:man} details the manipulability Jacobian before we use it to formulate the proposed Manipulability Motion Controller in Section \ref{sec:con}. Section \ref{sec:exp} describes our experimental setup and methodology. Finally, Section \ref{sec:res} details our experimental results and insights informed by the results.

\begin{figure*}[t]
    \centering
    \includegraphics[width=17cm]{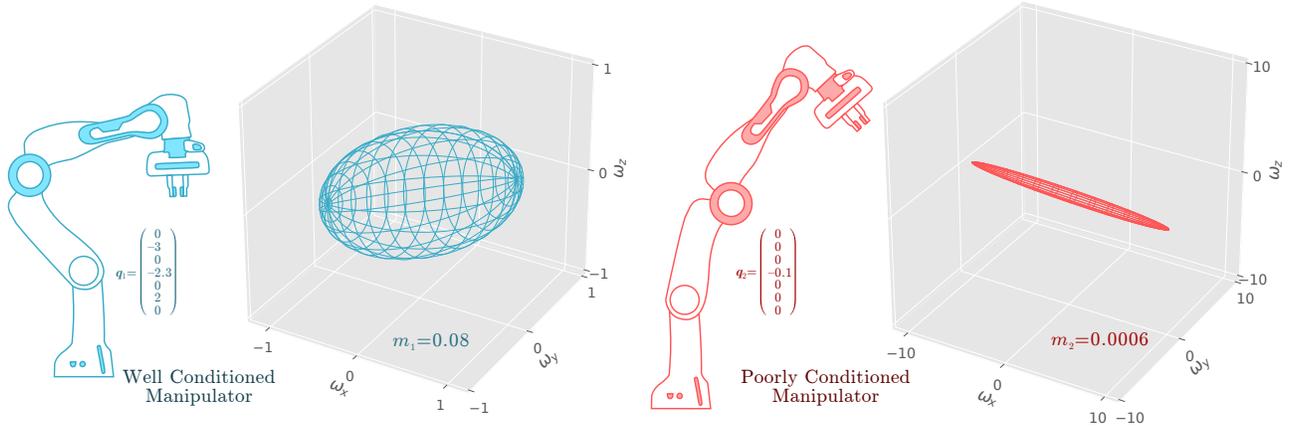}
    \caption{
        End-effector angular velocity ellipsoids created using the manipulator Jacobian at two different robot configurations $\vec{q}_1$ and $\vec{q}_2$, on a Panda robot. The ellipsoid depicts how easily the robot's end-effector can move with an arbitrary  angular velocity. The left ellipsoid shows the manipulator's configuration is well conditioned to rotate the end-effector in any direction. The right configuration is near singular as the end-effector will struggle to rotate around the y or z-axis. This ability to move is encapsulated in the manipulability denoted by $m_1$, and $m_2$.
    }
    \vspace{-14pt}
    \label{fig:large}
\end{figure*}

\section{Resolved-Rate Motion Control} \label{sec:rrmc}

The forward kinematics of a serial-link manipulator provides a non-linear surjective mapping between the joint space and Cartesian task space. This mapping is described as
\begin{equation} \label{m:rrmc0}
    \matfn{r}{t}=f(\vec{q}(t)) 
\end{equation}
where $\vec{q}(t) \in \R^n$ are the joint coordinates of the robot, $n$ is the number of joints, $\mat{r} \in \R^m$ is some parameterization of the end-effector pose, and the mapping function $f(\cdot)$ holds the geometrical information of the robot. The following derivations assume  the robot has a task space $\mathcal{T} \in \SE{3}$, and therefore $m = 6$. A redundant manipulator has a joint space dimension that exceeds the workspace dimension, i.e. $n > 6$. Taking the time derivative of (\ref{m:rrmc0})
\begin{equation}  \label{m:rrmc1}
    \vec{\nu}(t) = \matfn{J}{\vec{q}(t)} \dvec{q}(t) 
\end{equation}
where $\matfn{J}{\vec{q}_0} = \frac{\partial f(\vec{q})}{\partial \vec{q}} \big{|}_{\vec{q}=\vec{q}_0} \in \R^{6 \times n}$ is the manipulator Jacobian for the robot at configuration $\vec{q}_0$. Resolved-rate motion control is an algorithm which maps a Cartesian end-effector velocity $\vec{\nu}$ to the robot's joint velocity $\dvec{q}$. By rearranging (\ref{m:rrmc1}), the required joint velocities can be calculated as
\begin{equation}  \label{m:rrmc2}
    \dvec{q} = \matfn{J}{\vec{q}}^{-1} \ \vec{\nu}.
\end{equation}
Note that the $t$ variable in (\ref{m:rrmc2}) has been omitted for clarity. (\ref{m:rrmc2}) can only be solved when $\matfn{J}{\vec{q}}$ is square (and non-singular), which is when the robot has 6 degrees-of-freedom. 

For redundant robots there is no unique solution for (\ref{m:rrmc1}). Consequently, the most common solution is to use the Moore-Penrose pseudoinverse
\begin{equation} \label{m:rrmc3}
    \dvec{q} = \matfn{J}{\vec{q}}^{+} \ \vec{\nu}
\end{equation}
where the $(\cdot)^+$ denotes the pseudoinverse operation. The pseudoinverse will find $\vec{\nu}$ with the minimum Euclidean norm, which is useful for a real robot.

\section{Quadratic Programming} \label{sec:qp}

A general constrained quadratic programming (QP) problem is formulated as \cite{opt0}
\begin{align} \label{m:qp1}
    \min_x \quad f_o(\vec{x}) &= \frac{1}{2} \vec{x}^\top \mat{Q} \vec{x}+ \vec{c}^\top \vec{x}, \\ 
    \mbox{subject to} \quad \mat{A}_1 \vec{x} &= \vec{b}_1, \nonumber \\
    \mat{A}_2 \vec{x} &\leq \vec{b}_2, \nonumber \\
    \vec{d} &\leq  \vec{x} \leq \vec{e}. \nonumber 
\end{align}
where $f_o(\vec{x})$ is the objective function which is subject to the following equality and inequality constraints, and $\vec{d}$ and $\vec{e}$ represent the upper and lower bounds of $\vec{x}$. Typically, at least one constraint needs to be defined. A quadratic program is strictly convex when the matrix $\mat{Q}$ is positive definite \cite{opt0}.

Equation (\ref{m:rrmc3}) can be reformulated as a constrained quadratic programming problem in the form of (\ref{m:qp1})
\begin{align} \label{m:qp2}
    \min_{\dvec{q}} \quad f_o(\dvec{q}) &= \frac{1}{2} \dvec{q}^\top \mat{I}_n \dvec{q}, \\ 
    \mbox{subject to} \quad \matfn{J}{\vec{q}} \dvec{q} &= \vec{\nu} \nonumber
\end{align}
where $\mat{I}_{n}$ is an $n \times n$ identity matrix, and no inequality constraints need to be defined.
This optimisation minimises the control input, which in this case is joint velocities. 

\section{The Manipulability Jacobian} \label{sec:man}

\subsection{The Manipulability Measure}

A notable problem arises in serial-link manipulators when they approach a kinematic singularity. The manipulator Jacobian becomes ill-conditioned and the robot cannot move easily within its workspace and can cause required joint velocities to approach impossible levels \cite{peter}. At the singularity, the robot's task space is reduced by one or more degrees of freedom.

The manipulability measure in \cite{manip}, describes how well-conditioned the manipulator is to achieve an arbitrary velocity. It is a scalar
\begin{equation} \label{m:man1}
    m = \sqrt{ \mbox{det} \left(
        \matfn{J}{\vec{q}} \matfn{J}{\vec{q}}^\top
    \right) }
\end{equation}
which describes the volume of a 6-dimensional ellipsoid defined by 
\begin{equation} \label{m:man0}
    \matfn{J}{\vec{q}} \matfn{J}{\vec{q}}^\top.
\end{equation}

If this ellipsoid is close to spherical, then the manipulator can achieve any arbitrary end-effector velocity. A 6-dimensional ellipsoid is impossible to display, but the first three rows of the manipulator Jacobian represent the translational component of the end-effector velocity and the last three rows represent the end-effector angular velocity. Therefore, by using only the first or last three rows of a manipulator Jacobian in (\ref{m:man0}), the 3-dimensional translational or angular velocity ellipsoids respectively can be found and visualised. For example, Figure \ref{fig:large} show two angular velocity ellipsoids for two different robot configurations.

The ellipsoid can be described by three radii aligned with its principal axes. A small radius indicates the robot's inability to achieve a velocity in the corresponding direction. At a singularity, the ellipsoid's radius becomes zero along the corresponding axis and the volume becomes zero. If the manipulator's configuration is well conditioned, these ellipsoids will have a larger volume. The manipulability translational $m_t$ or rotational $m_r$ components can be found by taking the first, or last three rows of the manipulator Jacobian to calculate (\ref{m:man1}).

Manipulability is a favourable performance index for an optimisation function but it has a highly non-linear relationship with the manipulator's joint coordinates. Consequently, just as we use the manipulator Jacobian in (\ref{m:rrmc1}) to relate the joint velocities to the end-effector velocities, we can derive a manipulability Jacobian to relate the joint velocities to the rate of change of manipulability.

Taking the time derivative of (\ref{m:man1}), using the chain rule
\begin{align} \label{m:man2}
    \frac{\mathrm{d} \ m(t)}
         {\mathrm{d} t} = 
    \dfrac{1}
          {2m(t)} 
    \frac{\mathrm{d} \ \mbox{det} \left( \matfn{J}{\vec{q}} \matfn{J}{\vec{q}}^\top \right)}
         {\mathrm{d} t} 
\end{align}
we can write this \cite{ets} as
\begin{align} \label{m:man9}
    \dot{m}
    &=
    \vec{J}_m^\top \ \dvec{q}
\end{align}
where
\begin{equation} \label{m:man10}
    \vec{J}_m^\top
    =
    \begin{pmatrix}
        m \ \mbox{vec} \left( \mat{J} \mat{H_1}^\top \right)^\top 
        \mbox{vec} \left( (\mat{J}\mat{J}^\top)^{-1} \right) \\
        m \ \mbox{vec} \left( \mat{J} \mat{H_2}^\top \right)^\top 
        \mbox{vec} \left( (\mat{J}\mat{J}^\top)^{-1} \right) \\
        \vdots \\
        m \ \mbox{vec} \left( \mat{J} \mat{H_n}^\top \right)^\top 
        \mbox{vec} \left( (\mat{J}\mat{J}^\top)^{-1} \right) \\
    \end{pmatrix}
\end{equation}
is the manipulability Jacobian with $\vec{J}^\top_m \in \R^n$ and where the vector operation $\mbox{vec}(\cdot) : \R^{a \times b} \rightarrow \R^{ab}$ converts a matrix column-wise into a vector, and $\mat{H}_i \in \R^{6 \times n}$ is the $i^{th}$ component of the manipulator Hessian tensor $\mat{H} \in \R^{6 \times n \times n}$.

\section{Manipulability Motion Controller Design} \label{sec:con}

We use the manipulability Jacobian from (\ref{m:man10}) in our quadratic program. Recalling the general form of a quadratic program from (\ref{m:qp1}), the equation for the derivative of the manipulability in (\ref{m:man9}) fits the form of the linear component of the quadratic program. 
To prevent unreasonable or dangerous control inputs the control input is penalised \cite{opt0}. 
The final optimisation problem is
\begin{align} \label{m:qp3}
    \min_{\dvec{q}} \quad f_o(\dvec{q}) 
    &= 
    \frac{1}
         {2} 
    \dvec{q}^\top \lambda \mat{I}_n \dvec{q} - \vec{J}_m^\top \dvec{q}, \\ 
    \mbox{subject to} \quad \matfn{J}{\vec{q}} \dvec{q} &= \vec{\nu}. \nonumber
\end{align}
where $\lambda \in \mathbb{R}^+$ is a gain term, and we use $-\vec{J}_m$ to maximize rather than minimize manipulability. Since $\lambda \mat{I}_n$ is positive definite, the resulting optimisation problem is convex. The gain term $\lambda$ can be adjusted to tune how much the controller will minimise the control input relative to maximising the manipulability. If desired, an inequality constraint can be added to (\ref{m:qp3}) to bound the joint velocities
\begin{align*}
    \mbox{subject to} \quad \dvec{q}^- \leq \dvec{q} \leq \dvec{q}^+
\end{align*}
where $\dvec{q}^{-,+} \in \R^n$ are vectors representing the minimum and maximum joint velocity for each joint respectively.

We can force the optimiser to respect the joint position limits through velocity dampers and inequality constraints. Velocity dampers \cite{pp0} constrain velocities to prevent position limits from being exceeded. We form the velocity damper as
\begin{align} \label{m:qp4}
    \dot{q} \leq
    \eta
    \frac{\rho - \rho_s}
            {\rho_i - \rho_s}
    \qquad \mbox{if} \ \rho < \rho_i
\end{align}
where $\rho \in \mathbb{R}^+$ is the distance or angle to the nearest joint limit, $\eta \in \mathbb{R}^+$ is a gain which adjusts the aggressiveness of the damper, $\rho_i$ is the influence distance in which to activate the damper, and $\rho_s$ is the stopping distance in which the distance $\rho$ will never be able to reach or enter. We can incorporate (\ref{m:qp4}) into (\ref{m:qp3}) through an inequality contraint where rows are only added where $\rho < \rho_i$ for the respective joint. Assuming every joint is within the influence distance of the limit, the inequality constraint would be formed as
\begin{align}
    \mat{I}_n
    \dvec{q} &\leq
    \eta
    \begin{pmatrix}
        \frac{\rho_0 - \rho_s}
             {\rho_i - \rho_s} \\
        \vdots \\
        \frac{\rho_n - \rho_s}
             {\rho_i - \rho_s} 
    \end{pmatrix}.
\end{align}

However, the controller's primary task is to move the end-effector in a straight line to the desired pose, while exploiting the null-space of the differential kinematics to maximise manipulability. If the robot has no redundancy then there is no null-space to exploit. Therefore, we augment (\ref{m:qp3}) to incorporate slack \cite{opt1}. The slack is essentially intentional error, where the optimiser can choose to move components of the desired end-effector motion into the slack vector thereby deviating from the straight line motion. For both redundant and non-redundant robots, this means that the robot may stray from the straight line motion to improve manipulability and avoid a singularity, avoid running into joint position limits, or stay bounded by the joint velocity limits. We introduce slack into (\ref{m:qp3}) as
\begin{align} \label{m:qp5}
    \min_x \quad f_o(\vec{x}) &= \frac{1}{2} \vec{x}^\top \mathcal{Q} \vec{x}+ \mathcal{C}^\top \vec{x}, \\ 
    \mbox{subject to} \quad \mathcal{J} \vec{x} &= \vec{\nu}, \nonumber \\
    \mathcal{A} \vec{x} &\leq \mathcal{B}, \nonumber \\
    \vec{x}^- &\leq \vec{x} \leq \vec{x}^+ \nonumber
\end{align}

where
\begin{align}
    \vec{x} &= 
    \begin{pmatrix}
        \dvec{q} \\ \vec{\delta}
    \end{pmatrix} \in \mathbb{R}^{(n+6)}  \\
    \mathcal{Q} &=
    \begin{pmatrix}
        \lambda_q \mat{I}_{n \times n} & \mathbf{0}_{6 \times 6} \\ \mathbf{0}_{n \times n} & \lambda_\delta \mat{I}_{6 \times 6}
    \end{pmatrix} \in \mathbb{R}^{(n+6) \times (n+6)} \\
    \mathcal{J} &=
    \begin{pmatrix}
        \mat{J}(\vec{q}) & \mat{I}_{6 \times 6}
    \end{pmatrix} \in \mathbb{R}^{6 \times (n+6)} \label{eq:gain}
\end{align}
\begin{align}
    \mathcal{C} &= 
    \begin{pmatrix}
        \vec{J}_m \\ \mat{0}_{6 \times 1}
    \end{pmatrix} \in \mathbb{R}^{(n + 6)} \\
    \mathcal{A} &= 
    \begin{pmatrix}
        \mat{I}_{n \times n + 6} \\
    \end{pmatrix} \in \mathbb{R}^{(l + n) \times (n + 6)} \\
    \mathcal{B} &= \label{eq:gain2}
    \eta
    \begin{pmatrix}
        \frac{\rho_0 - \rho_s}
                {\rho_i - \rho_s} \\
        \vdots \\
        \frac{\rho_n - \rho_s}
                {\rho_i - \rho_s} 
    \end{pmatrix} \in \mathbb{R}^{n} \\
    \vec{x}^{-, +} &= 
    \begin{pmatrix}
        \dvec{q}^{-, +} \\
        \vec{\delta}^{-, +}
    \end{pmatrix} \in \mathbb{R}^{(n+6)}
\end{align}
and $\vec{\delta} \in \mathbb{R}^6$ is the added slack vector, and $\lambda_\delta \in \mathbb{R}^+$ is a gain term which adjusts the cost of the norm of the slack vector in the optimiser. The effect of this augmented optimisation problem is that the equality contraint is equivalent to
\begin{equation} \label{eq:qp2}
    \vec{\nu}(t) - \vec{\delta}(t) = \matfn{J}{\vec{q}} \dvec{q}(t)
\end{equation}
which clearly demonstrates the effect the slack $\vec{\delta}$ has on the end-effector velocity $\vec{\nu}$.

We have incorporated all components required to run the proposed controller in our open-source Robotics Toolbox for Python \cite{rtb}. The toolbox implements algorithms presented in \cite{ets} to calculate the manipulator Jacobian and Hessian for any manipulator whether the robot is modelled using DH, modified DH, URDF or ETS approaches. Furthermore, the toolbox can calculate the manipulability Jacobian and we present implementation details for the controller at the project website 
\href{https://jhavl.github.io/mmc}{jhavl.github.io/mmc}. 
We use the Python library $\mbox{qpsolvers}$ which implements the quadratic programming solver devised in \cite{quad} to optimise (\ref{m:qp5}). MMC can generally be solved in less than 3\unit{ms}, with speed improvements obtainable through multiprocessing.

\section{Experiments} \label{sec:exp}

\begin{table*}[t]
    \centering
    \renewcommand{\arraystretch}{1.3}

    \caption{Experiment 2: Results on 1000 Simulated PBS Tasks}
    \label{tab:results}

    \begin{tabular}{ c | M{3.1cm} | M{2.1cm} | M{2.2cm} | M{2.2cm} | M{2.2cm}}
    \hline
    {Robot} & Measure & {RRMC (Baseline)} & {Park \cite{gpm}} & {Baur \cite{gpm2}} & {MMC (ours)} \\
    \hline\hline
    Panda  & Mean Manipulability & 0.0693 & 0.0799, +15.3\% & 0.0785, +13.2\% & \textbf{0.0942, +35.9\%} \\
    & Mean Final Manipulability & 0.0692 & 0.0822, +18.4\% & 0.0801, +15.7\% & \textbf{0.0955, +38.0\%}\\
    & Failures & 12.4\% & 18.5\% & 12.7\% & \textbf{8.0\%} \\
    \hline
    UR-5   & Mean Manipulability & 0.0433 & 0.0433, +0.0\%  & 0.0433, +0.0\%  & \textbf{0.0563, +30.0\%} \\
    & Mean Final Manipulability & 0.0381 & 0.0381, +0.0\%  & 0.0381, +0.0\%  & \textbf{0.0460, +20.7\%}\\
    & Failures & 39.7\% & 39.7\% & 39.7\% & \textbf{25.6\%} \\
    \hline
    \end{tabular}
\end{table*}

We validate and evaluate our controller through testing on a real manipulator as well as in simulation on several different manipulators. We compare our Manipulability Motion Controller (MMC), as well as the motion controllers from Park \cite{gpm} and Baur \cite{gpm2}, to the standard Resolved-Rate Motion Controller (RRMC). RRMC is the baseline for standard reactive velocity control of a robot's end-effector. In each experiment, we choose a random initial joint configuration and a random end-effector goal pose. 
Each controller computes the spatial velocity to move from start to goal, and we capture the performance of the controllers for this motion.
Each controller finishes at the same desired end-effector pose but not necessarily the same joint configuration. 

The position-based servoing (PBS) scheme is 

\begin{equation} \label{m:pbs}
    \vec{\nu}_e = k \left( (\mat[0]{T}_e)^{-1} \bullet \mat[0]{T}_{e^*} \right)
\end{equation}
where $k$ is a gain term, $\mat[0]{T}_e \in \SE{3}$ is the end-effector pose in the robot's base frame, $\mat[0]{T}_{e^*} \in \SE{3}$ is the desired end-effector pose in the robot's base frame, and $\bullet$ represents composition. This scheme requests the robot's end-effector follow a straight path, in the robot's task space, to the goal pose.

For joint limit avoidance, we set $\eta = 1$, $\rho_i = 50\deg$, and $\rho_s = 2\deg$ in (\ref{eq:gain2}). Making $\eta$ larger increases the aggressiveness of the velocity damper.

We set $\lambda_q = 0.01$, and $\lambda_\delta = \frac{1}{e}$ in (\ref{eq:gain}) for all experiments, where $e$ represents the total error between the current and desired end-effector pose. We found that having $\lambda_q$ too high reduced the ability to maximize manipulability, while too low caused more extreme velocities within the robot leading to \emph{jerky} operation. 

There are many possible approaches to scheduling the slack penalty $\lambda_\delta$ along the trajectory.  The penalty should be low for as long as possible but be very high near the goal.
We found that setting the penalty to be inversely proportional to goal distance led to the optimiser having enough freedom to maximise manipulability along the trajectory, whilst ensuring that the goal is achieved. 
Too large a gain will limit the possible additional manipulability achievable, while too small a gain leads to the possibility that the slack will cancel out the desired velocity, leaving a large steady-state error. 

\subsection{Simulation Components}

For the simulated experiments, we use our open-source Robotics Toolbox for Python \cite{rtb}, and our simulator Swift \cite{swift} to simulate the Franka-Emika Panda (7 DoF), and the Universal Robot 5 (UR5, 6 DoF) manipulators. We reimplement the controllers devised by Park in \cite{gpm} and Baur in \cite{gpm2} where gains were set to be equivalent to ours (where possible) to ensure a fair comparison. Our implementations of these controllers are available at the project website.

\subsection{Experiment 2: Simulated Robots}

We compare our MMC to other reactive motion controllers in Park \cite{gpm} and Baur \cite{gpm2}, using the standard Resolved-Rate Motion Controller (RRMC) as a baseline. The controllers are compared by having them operate the PBS scheme in (\ref{m:pbs}) between 1000 randomly generated poses on each simulated serial-link manipulator. This experiment shows how much the MMC improves manipulability on average in a large-scale test, while also reporting failure numbers. We provide the results in Table \ref{tab:results}.

The initial configuration of the robot is generated by choosing random joint angles for each joint in the robot

\begin{equation} \label{m:rand}
    q_i = \mbox{rand} \left( q_{i, min} + 50\deg, \ q_{i, max} - 50\deg   \right)
\end{equation}
where $q_{i, min}$ and $q_{i, max}$ are the minimum and maximum valid joint angles (as specified by the manufacturer) for the joint $q_i$, the function $\mbox{rand}(a, b)$ returns a uniformly distributed number between $a$ and $b$, and the $50\deg$ offset is used to assist in keeping the configurations away from singular positions and self collisions. Configurations which result in self collisions are discarded.

The final pose is generated using (\ref{m:rand}) and using the forward kinematics of the robot to calculate the pose of the robot in that configuration. This pose is then used as $\mat[0]{T}_{e^*}$ in (\ref{m:pbs}).

\subsection{Physical Components}

For the physical experiments, we use our open-source Python library and ROS middleware to interface with the robot. We use the 7 degree-of-freedom Franka-Emika Panda robot in these experiments.

\subsection{Experiment 1: Physical Robot}

We compare MMC, Park \cite{gpm}, and RRMC by having them operate the PBS scheme in (\ref{m:pbs}) for several different scenarios. These scenarios reflect common operational situations which the controllers can encounter.
\renewcommand{\theenumi}{\alph{enumi}}
\begin{enumerate}
    \item The controllers servo between two poses in which the robot is well conditioned and not near a singularity. This reflects average and non-extreme operation of the robot. Furthermore, this is likely to be the most common scenario for a servo controller. 

    \item The controllers servo between two poses which differ greatly in orientation. In this experiment, the robot's end-effector starts facing the ground and finishes facing the sky. This reflects an extreme operation of the robot. 

    \item The controllers servo from a pose in which the robot is close to singularity and poorly conditioned to a pose in which the robot is well-conditioned. This experiment shows how each controller recovers the robot from a difficult pose. 

    \item The controllers servo from a pose in which the robot is well conditioned to a pose in which the robot is close to a singularity. In this experiment, the robot's final pose is on the outer bounds of the robot's task space. This experiment shows how the controllers behave when the robot moves towards a singular position. 
\end{enumerate}

\begin{figure*}[t!]
    \centering
    \begin{subfigure}{0.48\textwidth}
        \centering
        \includegraphics[width=1\textwidth, height=4.7cm]{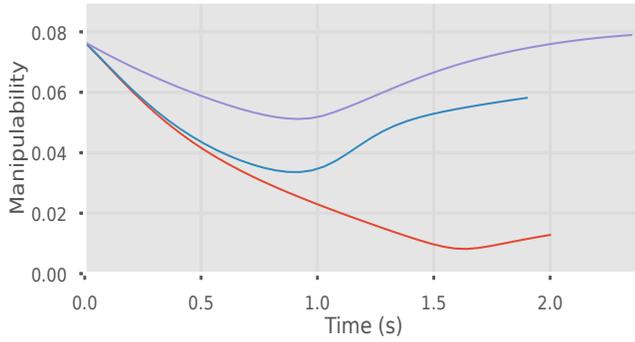}   
        \caption{Robot Manipulability Measure during a Normal Servo Operation}
        \label{fig:phys1}
    \end{subfigure}
    \hfill
    \begin{subfigure}{0.48\textwidth}
        \centering
        \includegraphics[width=1\textwidth, height=4.7cm]{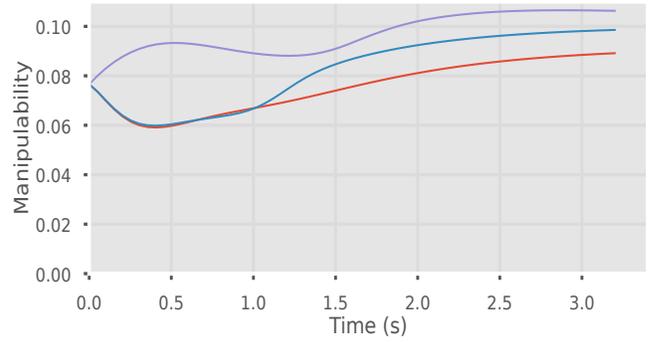}   
        \caption{Robot Manipulability Measure during a Complex Servo Operation}
        \label{fig:phys2}
    \end{subfigure}
    \\
    \begin{subfigure}{0.48\textwidth}
        \includegraphics[width=1\textwidth, height=4.7cm]{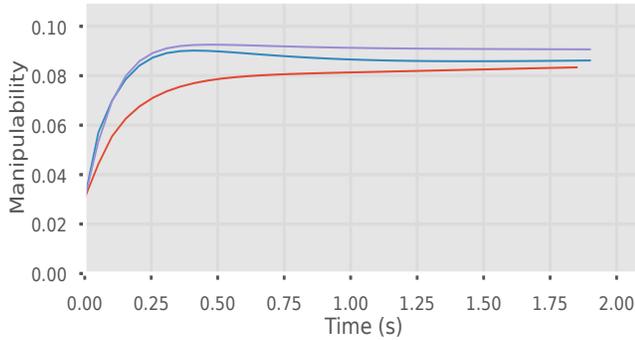}   
        \caption{Robot Manipulability Measure with the Robot's Initial Position near a Singularity}
        \label{fig:phys3}
    \end{subfigure}
    \hfill
    \begin{subfigure}{0.48\textwidth}
        \includegraphics[width=1\textwidth, height=4.7cm]{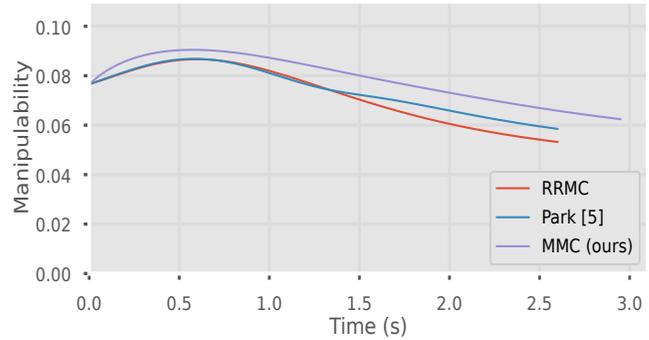}   
        \caption{Robot Manipulability Measure with the Robot's Final Pose near a Singularity}
        \label{fig:phys4}
    \end{subfigure}

    \caption{Experiment 1: Robot Manipulability Measure of RRMC and MMC during PBS in Various Scenarios}
\end{figure*}

\section{Results} \label{sec:res}

The average execution time of the MMC controller during the experiments was $2.53$\unit{ms} using an Intel i7-8700K CPU with 12 cores at 3.70GHz. The code is single threaded but execution time reduction is technically possible by multi-threading since the program is inherently parallelizable.

\begin{figure}[b!]
    \centering
    \includegraphics[height=5.0cm]{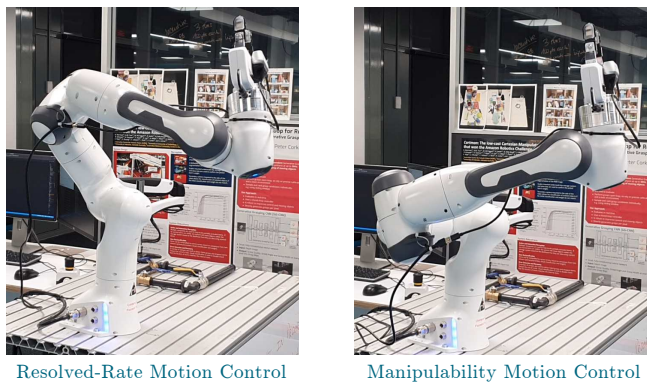}
    \caption{
        Final Pose of Experiment 1b: MMC provides a final pose which has 50\% better manipulability than RRMC.
    }
    \label{fig:practical}
\end{figure}

\subsection{Simulation Results}

Experiment 2 shows that the proposed MMC significantly improves the manipulability of a manipulator when compared to other controllers. 
The results, displayed in Table \ref{tab:results} show that on a Panda manipulator, MMC provides 35.9\% better mean manipulability and 38\% better final manipulability when compared to the baseline, and far exceeding the performance of the controllers from Park \cite{gpm} and Baur \cite{gpm2}. Additionally, we show that we can obtain 20.7\% and 30\% mean and final manipulability improvement respectively on the non-redundant UR5 manipulator where previous work provides no improvement.

MMC was also found to improve the robustness of the servo operation where 8\% of the 1000 Panda servo tasks failed, improving on the 12.4\% set by the baseline, and the other manipulability controllers proved worse. On the UR5 servo tasks, 25.6\% failed, greatly improving on the 39.7\% set by the baseline. The UR5 failures are high as the robot has a much smaller usable workspace due to having only 6 degrees-of-freedom. Failures reported by MMC were caused by the optimiser getting stuck in local minima -- it is impossible for the robot to run into joint limits, and due to the manipulability maximisation it is highly unlikely to run into a singularity. The other controllers mostly failed by attempting to exceed joint limits.
Avoiding, or detecting and exiting, local minima is an area we will pursue as future work. 

Over the 1000 servo tasks on the Panda, the average maximum deviation of MMC from the straight line motion, which the other 3 controllers follow exactly, was 108\unit{mm} while the average deviation was 66\unit{mm}. On the UR5, the average maximum deviation of MMC was 139\unit{mm} while the average deviation was 835\unit{mm}.
Deviation is the price we must pay to avoid joint position and velocity limits, achieve larger manipualbility gains, and have our controller operate on non-redundant robots. The acceptability of the deviation will depend on the application and can be adjusted through the $\lambda_\delta$ parameter.

\subsection{Physical Robot Results}

Figure \ref{fig:phys1} shows how the controller improves the manipulability during a normal servoing operation. This figure shows that MMC outperforms both Park and RRMC. This scenario reflects the most common operation of the robot and shows that MMC greatly improves manipulability on simple servoing tasks. The slack introduced by the end-effector velocity by MMC caused it to take longer than the other controllers.

Figure \ref{fig:phys2} shows that during a complex servoing operation, MMC maintains high manipulability. When the robot performs a complex movement, the other controllers which do not incorporate slack, can cause the robot to become \emph{twisted up}. This means that robots links are in close proximity and causes the robot to be poorly conditioned. MMC, as shown in Figure \ref{fig:phys2}, exhibits a high manipulability throughout the complex motion, while the controller from Park recovers some manipulability after the initial complex motion. Figure \ref{fig:practical}, displays the difference in final pose of MMC and RRMC for this experiment.

Figure \ref{fig:phys3} and \ref{fig:phys4} show edge cases of the robot recovering from a difficult configuration and entering a difficult configuration respectively. Figure \ref{fig:phys3} shows that both MMC and Park assists the robot in recovering from a poorly conditioned configuration much faster than RRMC, while MMC reaches a larger final manipulability. Figure \ref{fig:phys4}, shows the manipulability of the robot as it completes a reaching task where the final pose is on the outer bounds of the robot's task space. In this situation, despite the limited ability for slack to provide a benefit (due to the fully outstretched robot), MMC still clearly outperforms the other two controllers.

\section{Conclusions}
In this paper we have presented our Manipulability Motion Controller (MMC) as a purely reactive and robust controller for all manipulators.
Our results show MMC operating in several different scenarios on a real robot, and greatly improving the manipulability and robustness on a rigorous simulated experiment.
This translates to greatly improved usability when operating manipulators in a purely reactive manner.
We note that a limitation of this approach is the complexity around the mathematics in calculating the manipulability Jacobian, as well as the contained manipulator Hessian. Perhaps this is a reason that manipulability based controllers are not more widespread, despite being decades old. To mitigate this, we have incorporated all required tools required by MMC into our open-source Robotics Toolbox for Python \cite{rtb}, which can be used for DH, URDF, and ETS robot models.

\bibliographystyle{IEEEtran} 

\bibliography{ref} 

\end{document}